\documentclass[number,preprint,review,12pt]{elsarticle}

\usepackage{amssymb}
\usepackage{amsmath}
\usepackage{booktabs}
\usepackage{multirow}
\usepackage[hidelinks]{hyperref}
\usepackage{url}

\graphicspath{{./}{figures/}}

\begin{document}

\begin{frontmatter}

\title{SpermYOLO: A Coordinated YOLO-Based Detector for Accurate and Efficient Sperm and Impurity Detection in Microscopic Images}

\author[affilA]{Shengqi Chen\fnref{equal}}
\ead{shengqichen@bupt.edu.cn}
\author[affilA]{Zilin Wang\fnref{equal}}
\ead{roff972165375@163.com}
\author[affilA]{Xingyu Pan}
\ead{2023210693@bupt.cn}
\author[affilB]{Wenting Yu}
\ead{amberyoui@bupt.edu.cn}
\author[affilA]{Pengchao Deng}
\ead{202111070202@sdust.edu.cn}
\author[affilA]{Guohua Wu\corref{cor1}}

\cortext[cor1]{Corresponding author}
\ead{wuguohua@bupt.edu.cn}
\fntext[equal]{These authors contributed equally.}

\address[affilA]{School of Electronic Engineering, Beijing University of Posts and Telecommunications, 10 Xitucheng Road, Haidian District, Beijing 100876, China}
\address[affilB]{School of Information and Communication Engineering, Zhongyuan University of Technology, 41 Zhongyuan Middle Road, Zhengzhou, Henan Province 450007, China}

\begin{abstract}
Accurate sperm detection is essential for computer-assisted semen analysis, yet it remains challenging in microscopic images due to dense distributions, visually similar artifacts, and sperm-like impurities.
In this paper, we propose SpermYOLO, a coordinated and compact YOLOv11-derived framework for joint sperm and impurity detection in microscopic images.
SpermYOLO introduces four architectural improvements: C3k2-IDB for channel-wise discriminative feature extraction, D2SEM for spatial--spectral semantic enhancement, MFM for adaptive multi-scale feature fusion, and the DESD Head for detail-enhanced shared prediction.
Experiments on the SVIA semen microscopic imaging benchmark show that SpermYOLO achieves 97.2\% sperm AP and 75.4\% impurity AP, outperforming generic detectors, dedicated sperm detection models, and improved YOLO variants.
Compared with the baseline model, SpermYOLO improves sperm AP, impurity AP, $\mathrm{mAP}_{50}$, and $\mathrm{mAP}_{50:95}$ by 1.6, 10.0, 5.8, and 2.7 percentage points, respectively, while preserving a lightweight model scale.
Cross-scene evaluation on the SDTB testicular-biopsy microscopy benchmark shows that SpermYOLO remains effective with extremely small sperm targets and complex tissue backgrounds, achieving the highest $\mathrm{mAP}_{50}$ and $\mathrm{mAP}_{50:95}$ of 74.8\% and 31.2\%, respectively.
Ablation studies and qualitative analyses further support these improvements by demonstrating the contributions of the proposed modules and showing more focused feature response patterns than the baseline model.
These findings suggest that SpermYOLO is an effective and efficient approach for sperm detection in challenging microscopic imaging scenarios.
\end{abstract}


\end{frontmatter}


\section{Introduction}

Infertility has emerged as a major public health concern, affecting approximately 15\% of the population worldwide~\citep{sang2023understanding}. Accumulating evidence indicates that male-related causes contribute to 40--50\% of infertility cases~\citep{agarwal2021male,kumar2015trends}, and global trends point to a concerning decline in semen quality among young men in recent decades~\citep{luo2023global}. In this context, semen analysis, which assesses key sperm parameters including motility~\citep{franken2012semen}, morphology, and sperm concentration~\citep{gatimel2017sperm}, has become a cornerstone for the clinical evaluation of male-factor infertility~\citep{agarwal2021male,boitrelle2021sixth}.

Traditional semen analysis relies primarily on manual microscopic examination, which is time-consuming, labor-intensive, and subject to inter-operator variability~\citep{daoud2016inter}. To address these issues, computer-assisted sperm analysis (CASA) systems have been developed to improve objectivity and efficiency through automated image-based analysis~\citep{ates2021integrated}. As a core component of CASA, sperm detection identifies and localizes sperm cells in microscopic images, forming the basis for subsequent parameter quantification~\citep{elsayed2015development}. 
Early sperm detection employed classical image processing algorithms, including threshold segmentation~\citep{urbano2017automatic,abbiramy2010spermatozoa} and ellipse detection~\citep{yang2014head,mahdavi2011sperm}. While these approaches have demonstrated initial utility, their reliance on manual priors, handcrafted features, and low-level visual cues fundamentally limits their robustness in sperm microscopy images~\citep{mortimer2015future,zhao2020survey}.

In recent years, deep learning (DL) has achieved remarkable progress in medical image analysis~\citep{zhou2021review}, with superior performance reported across disease classification~\citep{chen2025review}, lesion segmentation~\citep{rayed2024deep}, and cell detection~\citep{huo2025survey}, demonstrating its promise for sperm detection. 
Conventional DL-based object detection models can be categorized into two major paradigms. Two-stage models exemplified by Faster R-CNN~\citep{ren2017faster} first generate candidate regions and then perform classification and bounding box regression on these proposals, leading to high computational complexity and limited real-time performance~\citep{lamichhane2025cnn}. 
One-stage detectors---including SSD~\citep{liu2016ssd} and YOLO variants~\citep{li2022yolov6,varghese2024yolov8,wang2024yolov10,khanam2024yolov11,tian2025yolov12,lei2025yolov13}---instead predict classes and boxes in a single forward pass, achieving higher detection efficiency. 
In parallel, Transformer-based detectors (DETR~\citep{carion2020end} and RT-DETR~\citep{zhao2024detrs}) represent an alternative paradigm that leverages self-attention to enable long-range dependency modeling and end-to-end detection but incurs higher computational cost that limits cost-effective deployment~\citep{li2023transformer}.

Several studies have applied generic object detectors directly to sperm detection~\citep{wu2021preliminary,dobrovolny2023study,yuzkat2023detection,chen2022svia} and demonstrated preliminary feasibility.
However, their reported performance remained limited, indicating insufficient adaptation to sperm microscopy.
In particular, dense distributions with frequent overlaps, grayscale-similar artifacts, and sperm-like impurities pose substantial challenges to reliable localization and discrimination.
Although specialized sperm detection models have been developed to enhance sperm detection through small-object feature preservation, spatial detail enhancement, multi-scale fusion, and contextual modeling~\citep{hidayatullah2021deepsperm,zou2022tod,zhang2023todnet,zhu2023yolov5s,li2025yolov8sta,pei2025multi,zhang2025spermd}, robust detection remains challenging, especially for joint sperm--impurity detection under complex microscopic conditions~\citep{chen2024active,meng2025yolov7}.

Recent work has highlighted the importance of incorporating task-specific improvements into a coordinated paradigm for sperm microscopy~\citep{zhang2025spermd}.
Building upon this, a lightweight detector with complementary adaptations is needed for accurate and efficient sperm--impurity detection.
We therefore present SpermYOLO, a compact YOLOv11-derived one-stage detector designed to improve sperm detection in microscopic images while explicitly recognizing interfering impurities.
To this end, SpermYOLO adapts the detection pipeline across feature extraction, semantic enhancement, multi-scale fusion, and prediction.
Specifically, C3k2-IDB replaces the C3k2 modules with an inverted bottleneck and lightweight channel recalibration to strengthen channel-wise feature discrimination.
The Dual-Domain Spectral Enhancement Module (D2SEM) incorporates spatial--spectral semantic modeling at the deepest feature stage to enhance high-level semantic representation.
The Multi-Scale Feature Fusion Module (MFM) replaces rigid concatenation with adaptive scale-wise fusion in the neck to aggregate features across resolutions.
The Detail-Enhanced Shared Detection Head (DESD Head) integrates detail-enhanced convolution with cross-scale parameter sharing for fine-grained localization and reduced head redundancy.

\begin{figure}[ht]
\centering
\includegraphics[width=0.85\textwidth]{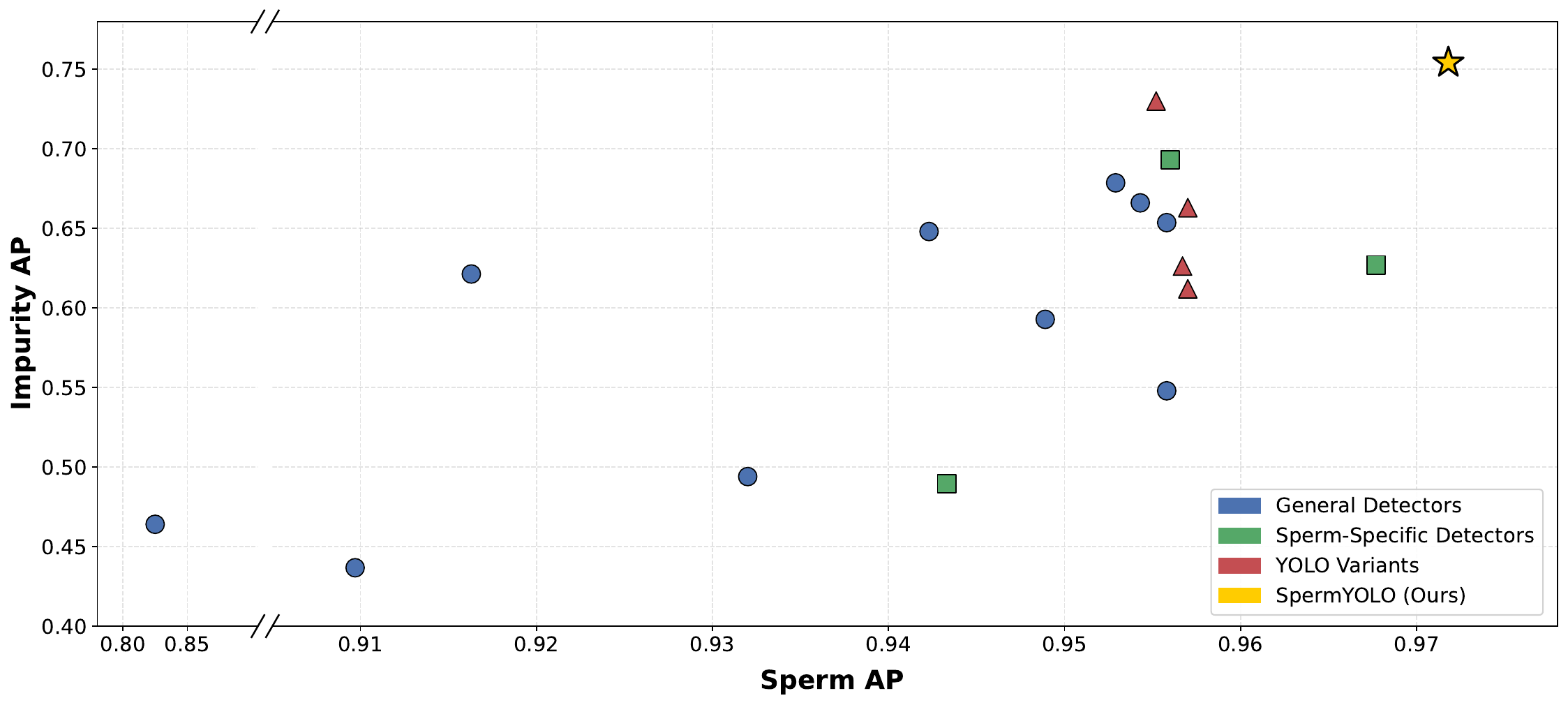}
\caption{Sperm AP vs.\ Impurity AP on the SVIA dataset. SpermYOLO achieves the best trade-off between the two classes, outperforming generic detectors, dedicated sperm detection models, and YOLO variants.}
\label{fig:sperm_vs_impurity}
\end{figure}

Experiments on the semen microscopic imaging benchmark SVIA~\citep{chen2022svia} demonstrate that SpermYOLO achieves the highest accuracy for joint sperm--impurity detection, outperforming generic detectors, dedicated sperm detection models, and improved YOLO variants while maintaining high efficiency and a compact model architecture.
As shown in Fig.~\ref{fig:sperm_vs_impurity}, SpermYOLO occupies the upper-right region of the plane, indicating strong performance on both classes; in contrast, existing sperm-specific detectors and YOLO variants do not achieve the best performance on both classes simultaneously.
Additional evaluation on the SDTB~\citep{zhang2025spermd} benchmark further shows that SpermYOLO remains effective in testicular-biopsy microscopy with extremely small sperm targets and complex tissue backgrounds.
The primary contributions of this study are summarized as follows:
\begin{enumerate}
  \item We propose SpermYOLO, a YOLOv11-derived detector built on four coordinated architectural improvements: C3k2-IDB, D2SEM, MFM, and the DESD Head. These components target feature discrimination, semantic representation, multi-scale fusion, and fine-grained prediction, with their effectiveness supported by ablation experiments.
  \item We evaluate SpermYOLO on the SVIA benchmark for semen microscopic imaging. 
  SpermYOLO achieves the best joint sperm--impurity detection accuracy among the evaluated generic detectors, dedicated sperm detection models, and improved YOLO variants, while maintaining high efficiency and a compact model architecture.
  \item We conduct cross-scene evaluation on the SDTB testicular-biopsy microscopy benchmark, showing that the proposed method remains effective with extremely small sperm targets and complex tissue backgrounds.
  \item We provide qualitative analyses of detection results and heatmaps, illustrating SpermYOLO's detection behavior and feature response patterns relative to the baseline model.
\end{enumerate}

The remainder of this paper is organized as follows. 
\hyperref[sec:related]{Section~\ref*{sec:related}} reviews related work. 
\hyperref[sec:methodology]{Section~\ref*{sec:methodology}} details the SpermYOLO architecture and proposed modules. 
\hyperref[sec:experiments]{Section~\ref*{sec:experiments}} describes the datasets, evaluation metrics, and implementation details. 
\hyperref[sec:results]{Section~\ref*{sec:results}} presents experimental results and analysis. 
Finally, \hyperref[sec:conclusion]{Section~\ref*{sec:conclusion}} concludes the paper.

\section{Related Work}
\label{sec:related}

\subsection{Traditional Methods}

In early sperm detection, Abbiramy et al.~\citep{abbiramy2010spermatozoa} employed gradient-based threshold segmentation combined with morphological operations for sperm counting, achieving 93\% accuracy. 
Urbano et al.~\citep{urbano2017automatic} detected sperm heads using filter-based feature enhancement and adaptive threshold segmentation, reporting approximately 95\% detection accuracy with false-positive rates below 1\%. 
Yang et al.~\citep{yang2014head} modeled sperm heads as elliptical objects and applied an improved Multiple Birth and Cut (MBC) algorithm, yielding 88\% precision and 79\% recall. 
Mahdavi et al.~\citep{mahdavi2011sperm} achieved over 94\% accuracy by performing elliptical detection with morphological filtering. 
Despite these promising results under controlled conditions, these methods rely heavily on manually designed priors and handcrafted features, leading to limited robustness~\citep{abbiramy2010spermatozoa}, particularly in dense sperm distributions with frequent occlusions and overlaps~\citep{urbano2017automatic,mahdavi2011sperm}. Moreover, computationally intensive image processing restricts real-time applicability~\citep{urbano2017automatic,yang2014head}.

\subsection{General Object Detection Models}

Wu et al.~\citep{wu2021preliminary} employed SSD on a self-constructed testicular biopsy dataset and achieved an $\mathrm{mAP}_{50}$ of 74.1\%, demonstrating the feasibility of DL-based sperm detection, but the low recall (37.6\%) and poor recognition of morphologically abnormal spermatozoa exposed clear limitations. 
Similar issues were observed when YOLOv5 was applied to an annotated VISEM subset by Dobrovolny et al.~\citep{dobrovolny2023study}, where the model achieved an mAP of 72.15\% but remained sensitive to grayscale-similar artifacts and showed limited recall.
Yuzkat et al.~\citep{yuzkat2023detection} benchmarked YOLOv5, SSD, EfficientDet, Detectron2, and Mask R-CNN on SeSViD, with YOLOv5 obtaining the best $\mathrm{mAP}_{50}$ of 88\%. However, performance still degraded under dense distributions and low-quality microscopy images, with a noticeable class-wise disparity favoring non-sperm structures. 
Chen et al.~\citep{chen2022svia} constructed the large-scale SVIA dataset and evaluated SSD, YOLOv3/v4, and Faster R-CNN on its detection subset (Subset-A). Their results revealed that off-the-shelf detectors remain insufficiently accurate and robust in complex scenarios containing impurities.
Collectively, these studies indicate that generic detectors lack sufficient adaptation to sperm-specific visual patterns and complex microscopic imaging conditions.

\subsection{Specialized Sperm Detection Models}

The performance gap of generic detectors has spurred the development of specialized architectures tailored for sperm microscopy. 
Previous studies mainly focused on preserving small-object information under dense, low-contrast, and blurred conditions.
DeepSperm~\citep{hidayatullah2021deepsperm} features a YOLO-based pipeline with dropout regularization and a single small-object detection layer for real-time bull sperm detection. 
TOD-CNN~\citep{zou2022tod} introduced cross-layer feature transfer and multi-scale fusion to retain local information for tiny sperm. 
TOD-Net~\citep{zhang2023todnet} further incorporated self-attention to strengthen feature interaction for small-object localization. 
These studies established that sperm detection benefits from task-specific spatial feature preservation. 
Nevertheless, their failure modes, including blurred sperm, edge targets, and sperm-like impurities, indicate that spatial detail preservation alone does not fully resolve fine-grained discrimination.

Recent sperm detectors have further refined this strategy within more efficient one-stage frameworks. 
YOLOv5s-SA~\citep{zhu2023yolov5s} combines depthwise separable convolution, enhanced multi-scale feature fusion, and shuffle attention to strengthen fine-grained representation with reduced computational cost.
YOLOv8-STA~\citep{li2025yolov8sta} integrates SPD-Conv, triplet attention, and an additional small-object detection head to mitigate information loss during downsampling.
Pei et al.~\citep{pei2025multi} partitioned microscope images into overlapping sub-images to preserve spatial details during input resizing.
However, these methods remain largely centered on spatial-domain detail retention, leaving the interference from visually similar impurities insufficiently addressed.

Focusing on joint sperm--impurity detection, Chen et al.~\citep{chen2024active} proposed ACTIVE, a dual-branch feature extraction network with a cross-conjugate feature pyramid network to enhance tiny-object features and fuse positional and semantic information. 
Meng et al.~\citep{meng2025yolov7} improved YOLOv7 using switchable atrous convolution and Wise-IoU loss to enlarge the receptive field and improve robustness to scale variation. 
Nonetheless, the inter-class performance gap on SVIA Subset-A indicates that impurities remain substantially harder to detect than sperm, underscoring the need for more discriminative representations beyond stronger feature extraction.

Most closely related to our work, Zhang et al.~\citep{zhang2025spermd} link detector-stage design with sperm-specific failure modes by identifying three challenges in sperm microscopy: visual similarity between sperm and background noise, insufficient local contextual modeling, and diminishment of tiny sperm features during feature fusion.
They subsequently developed SpermDet, which integrates structure-aware alignment fusion, local context enhancement, and semantic-aware dual-path fusion, reflecting a broader trend toward coordinated task-specific adaptation in sperm detection.
However, despite its strong performance across multiple benchmarks, further exploration of sperm detectors that balance task-specific architectural design and computational efficiency remains important for CASA systems, where detection serves as a front-end step for subsequent sperm analysis.

Inspired by the above studies, we develop SpermYOLO as a compact one-stage detector for joint sperm--impurity detection that incorporates targeted architectural improvements across feature extraction, semantic enhancement, multi-scale fusion, and prediction to improve detection performance while maintaining computational efficiency.

\section{Methodology}
\label{sec:methodology}

SpermYOLO adopts YOLOv11~\citep{khanam2024yolov11} as the foundation for its efficient one-stage architecture combining convolutional feature extraction, multi-scale feature aggregation, and decoupled prediction heads. 
Considering the real-time requirements of microscopic analysis, the nano variant (YOLOv11n) is selected as the baseline, balancing computational efficiency and detection performance while providing a compact design for targeted modifications.

In the original YOLOv11 architecture, C3k2 blocks serve as the primary feature extraction units, C2PSA refines high-level semantic representations, the neck aggregates multi-scale features, and the detection head generates localization and classification predictions.
However, this general-purpose architecture is not specifically tailored to sperm microscopy, where small, low-contrast sperm heads, visually similar impurities, and progressive loss of discriminative details during downsampling and fusion hinder reliable detection.
SpermYOLO therefore introduces targeted modifications across the feature extraction, semantic enhancement, multi-scale fusion, and prediction stages to improve discriminative representation and fine-grained localization while maintaining model compactness.

\subsection{Overall Architecture}

\begin{figure}[t]
\centering
\includegraphics[width=0.95\textwidth]{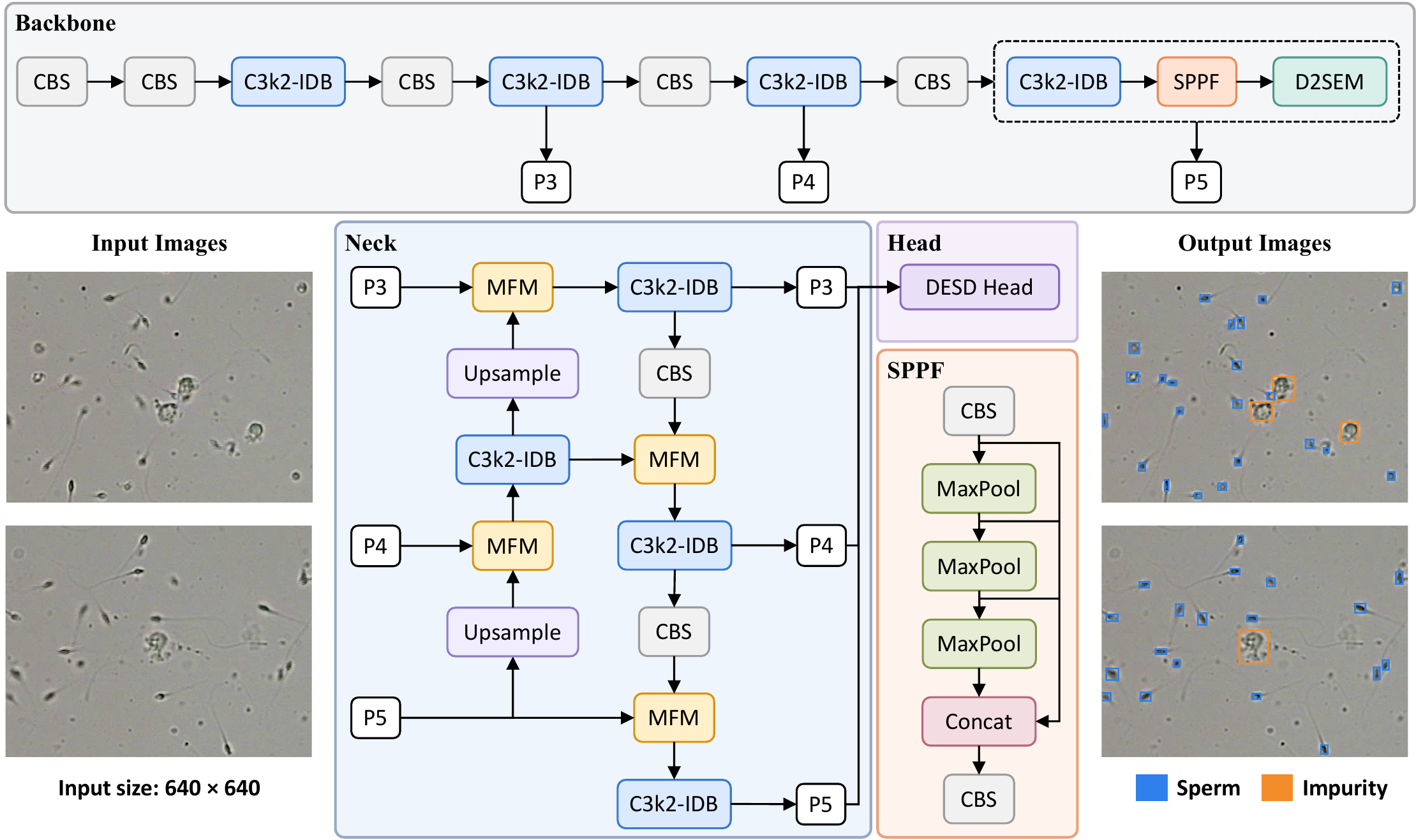}
\caption{Overall Architecture of SpermYOLO}
\label{fig:architecture}
\end{figure}

As shown in Fig.~\ref{fig:architecture}, SpermYOLO adapts YOLOv11 through four complementary stages. 
C3k2-IDB replaces the C3k2 modules with an inverted bottleneck and lightweight channel recalibration.
D2SEM replaces the C2PSA module after spatial pyramid pooling and enhances high-level semantic representations through spatial--spectral feature modeling.
MFM replaces concatenation-based fusion in the neck with adaptive scale weighting. 
The DESD Head applies shared detail-enhanced convolutions to the multi-scale detection features. 
Together, these modifications form a coordinated adaptation of different detection stages rather than independent architectural add-ons.

\subsection{C3k2-IDB: Inverted Discriminative Bottleneck}

\begin{figure}[t]
\centering
\includegraphics[width=0.86\textwidth]{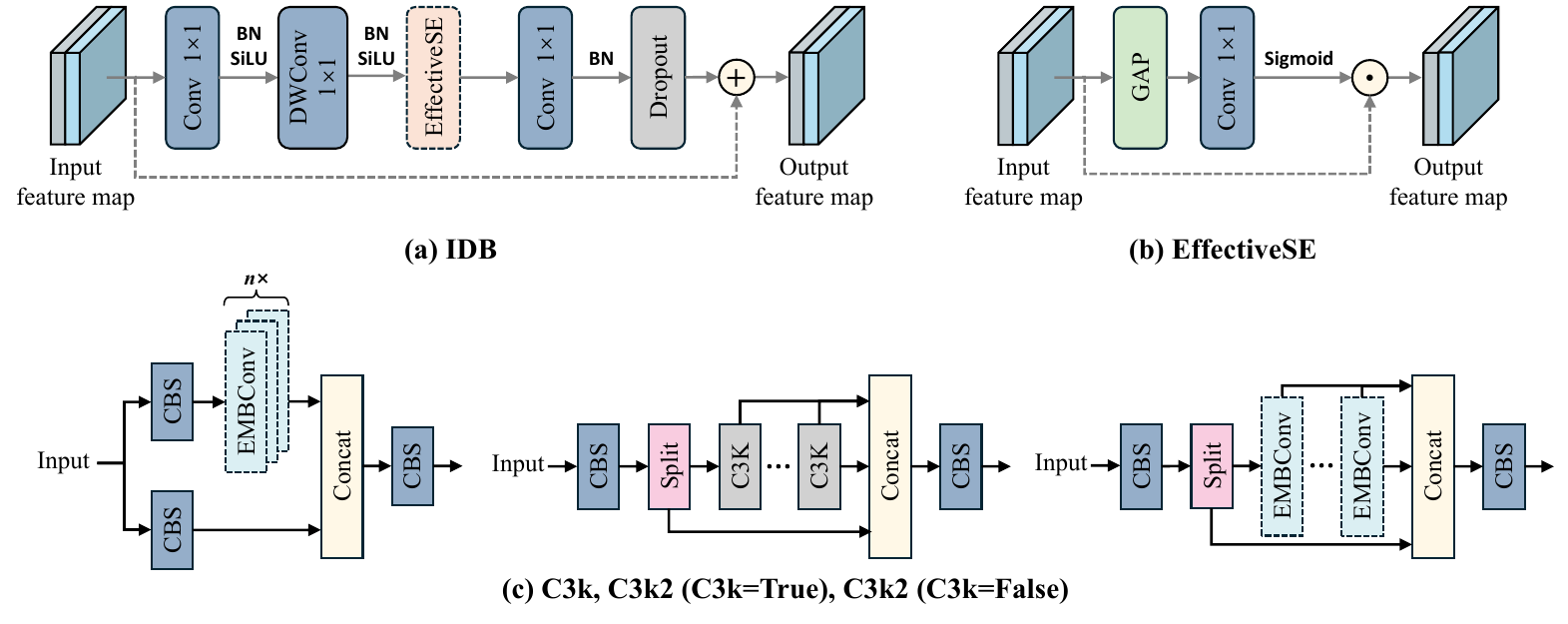}
\caption{Structure of the C3k2-IDB Module}
\label{fig:c3k2_idb}
\end{figure}

The C3k2 module is the primary convolutional feature extraction unit in YOLOv11.
While its standard bottleneck design is efficient, there is no explicit recalibration of channel responses to emphasize discriminative features and suppress background-related responses.
This limitation is particularly relevant for sperm microscopy because weak sperm-head contours and small impurity boundaries can blend with low-contrast background structures.
SpermYOLO therefore replaces the C3k2 modules with C3k2-IDB, converting the conventional local bottleneck into an inverted, channel-recalibrated block, as shown in Fig.~\ref{fig:c3k2_idb}.

The core unit in C3k2-IDB follows the mobile inverted bottleneck convolution (MBConv) design~\citep{tan2019efficientnet} and incorporates effective squeeze-and-excitation (EffectiveSE)~\citep{lee2020centermask}.
For an input feature map $x$, the block first expands the channel dimension using a $1\times1$ pointwise convolution.
It then applies a $3\times3$ depthwise convolution (DWConv) for spatial mixing and recalibrates the expanded features using EffectiveSE. 
The result is projected through another $1\times1$ pointwise convolution and a residual connection is retained to preserve the input representation.
The entire process is written as follows:
\begin{equation}
\mathcal{F}_{\mathrm{IDB}}(x)=x+
\mathrm{Proj}\left(
\mathrm{EffectiveSE}\left(
\mathrm{DWConv}_{3\times3}\left(
\mathrm{Expand}(x)
\right)\right)\right)
\label{eq:idb_transform}
\end{equation}

Here, $\mathrm{Expand}(\cdot)$ and $\mathrm{Proj}(\cdot)$ denote channel expansion and projection via $1\times1$ convolutions, respectively; dropout is applied after projection only during training.

EffectiveSE recalibrates channels using a global descriptor without the reduction-and-expansion multilayer perceptron adopted in conventional squeeze-and-excitation blocks.
Given an intermediate feature map $u$, it applies a $1\times1$ convolution to the pooled descriptor and uses HardSigmoid~\citep{howard2019mobilenetv3} to produce channel-wise weights as follows:
\begin{equation}
\mathrm{EffectiveSE}(u)=u\odot
\mathrm{HardSigmoid}\left(
\mathrm{Conv}_{1\times1}\left(\mathrm{GAP}(u)\right)
\right)
\label{eq:effective_se}
\end{equation}

where $\mathrm{GAP}(\cdot)$ denotes global average pooling over the spatial dimensions, and $\odot$ denotes element-wise multiplication.
This design keeps full channel resolution during attention computation and avoids an additional channel bottleneck. In SpermYOLO, C3k2-IDB is designed to enhance channel-wise feature discrimination for sperm and impurity cues while maintaining lightweight computation.

\subsection{D2SEM: Dual-Domain Spectral Enhancement Module}

In YOLOv11, C2PSA refines high-level semantic features by modeling long-range interactions with partial self-attention.
However, the feed-forward branch in the standard partial self-attention (PSA) block uses pointwise convolutions and remains confined to spatial-domain feature transformation.
In sperm microscopy, weak sperm-head contrast and visual similarity between impurities and background artifacts may involve both local texture and frequency-related characteristics.
Relying only on spatial-domain channel mixing may therefore underuse spectral cues in deep semantic features.

D2SEM is constructed by replacing the pointwise-convolution feed-forward network inside each PSA block of C2PSA with the spectral-enhanced feed-forward network (SEFFN) design from TransMamba~\citep{sun2024transmamba}, forming a spectral-enhanced PSA (SEPSA) block.
In this way, D2SEM preserves C2PSA's self-attention branch while augmenting its feed-forward branch with lightweight frequency-domain channel modulation.

\begin{figure}[t]
\centering
\includegraphics[width=0.86\textwidth]{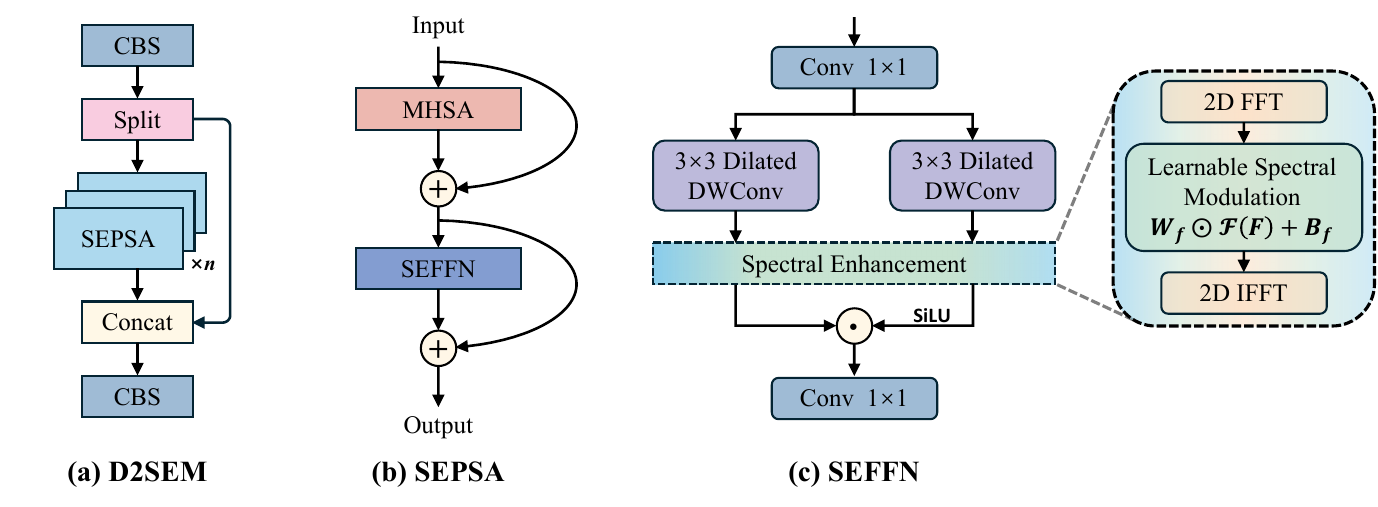}
\caption{Structure of the D2SEM Module}
\label{fig:d2sem}
\end{figure}

As shown in Fig.~\ref{fig:d2sem}(a), D2SEM retains the split-and-fuse structure of C2PSA.
A $1\times1$ projection splits the input into a bypass branch and a branch processed by stacked SEPSA blocks; their outputs are then concatenated and fused by a final $1\times1$ projection.

Figure~\ref{fig:d2sem}(b) details SEPSA, where the multi-head self-attention (MHSA) branch follows the original PSA design and SEFFN replaces the feed-forward update, with residual connections applied to both branches.
Following Fig.~\ref{fig:d2sem}(c), given an input feature map $F\in\mathbb{R}^{C\times H\times W}$, SEFFN first expands the channel dimension to $2C_h$ via a $1\times1$ projection, where $C_h=rC$ and $r$ denotes the expansion ratio.
The expanded feature is divided into two branches, $F_1$ and $F_2$, which undergo $3\times3$ dilated DWConv processing before spectral enhancement.
Each branch is transformed by a two-dimensional fast Fourier transform (2D FFT) and modulated by learnable spectral-domain weights and biases:
\begin{equation}
\hat{F}_i = W_{fi} \odot \mathcal{F}(F_i) + B_{fi},\quad i\in\{1,2\}
\label{eq:d2sem_spectral}
\end{equation}
where $\mathcal{F}(\cdot)$ denotes the 2D FFT, $\odot$ denotes element-wise multiplication, and $W_{fi}$ and $B_{fi}$ are learnable spectral-domain weight and bias terms broadcast to the corresponding frequency-domain feature.

The inverse 2D FFT, $\mathcal{F}^{-1}(\cdot)$, then maps both enhanced branches back to the spatial domain.
SEFFN applies SiLU~\citep{elfwing2018sigmoid} to the restored second branch and multiplies it with the restored first branch before a final $1\times1$ projection:
\begin{equation}
\mathrm{SEFFN}(F)=
\mathrm{Proj}_{\mathrm{out}}\left(
\mathrm{SiLU}(\mathcal{F}^{-1}(\hat{F}_2))\odot
\mathcal{F}^{-1}(\hat{F}_1)
\right)
\label{eq:seffn_output}
\end{equation}

Overall, D2SEM extends C2PSA to joint spatial--spectral semantic refinement, thereby introducing frequency-domain channel cues for low-contrast sperm and impurity discrimination.

\subsection{MFM: Multi-Scale Feature Fusion Module}

In the neck, YOLOv11 aggregates multi-scale features by concatenating spatially aligned feature maps from adjacent stages.
This fixed fusion combines incoming scales but does not explicitly estimate their channel-wise contributions.
For sperm microscopy, the useful scale can differ by feature channel: fine-resolution maps help localize small sperm-head contours and impurity boundaries, whereas coarser maps carry context for distinguishing sperm from impurities and background clutter.

\begin{figure}[t]
\centering
\includegraphics[width=0.86\textwidth]{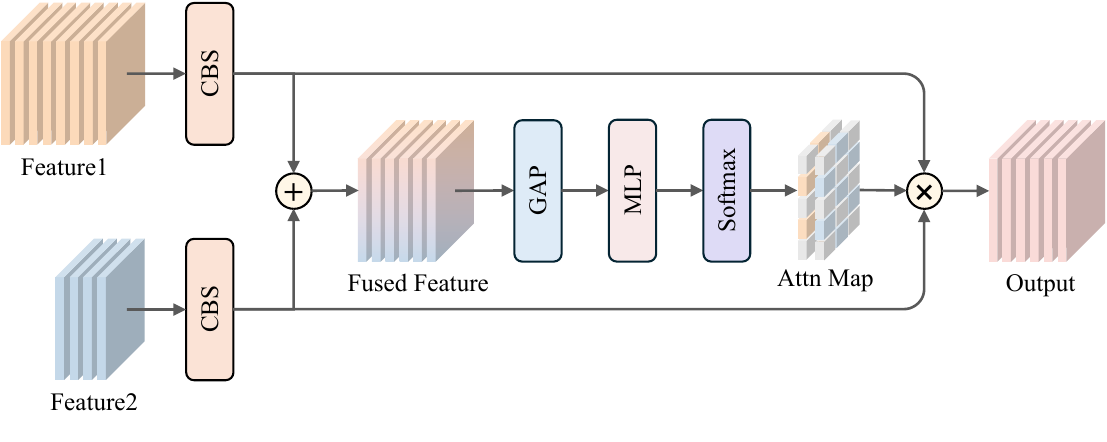}
\caption{Structure of the MFM Module}
\label{fig:mfm}
\end{figure}

Motivated by the multi-scale feature modulation strategy in DCMPNet~\citep{zhang2024dcmpnet}, SpermYOLO adopts MFM in the neck to introduce adaptive scale-wise aggregation, as shown in Fig.~\ref{fig:mfm}.
Given two spatially aligned feature maps $F_1$ and $F_2$ to be fused in the neck, SpermYOLO adapts vanilla MFM with lightweight channel-alignment projections that map them to a common dimension $C_m$:
\begin{equation}
f_k = \phi_k(F_k), \quad k\in\{1,2\}
\label{eq:mfm_projection}
\end{equation}

where $\phi_k$ denotes a $1\times1$ projection.
The aligned features are summed to form the attention input.
Global average pooling then compresses this sum, and a two-layer MLP produces scale-specific channel scores.
A softmax over the two input scales converts these scores into channel weights:

\begin{equation}
\alpha =
\mathrm{Softmax}\left\{
\mathrm{MLP}\left[
\mathrm{GAP}\left(\sum_{k=1}^{2}f_k\right)
\right]\right\}
\label{eq:mfm_attention}
\end{equation}

The final fused feature is the channel-wise weighted sum of the aligned inputs:

\begin{equation}
\mathrm{MFM}(f_1,f_2)=
\sum_{k=1}^{2}\alpha_k\odot f_k 
\label{eq:mfm_output}
\end{equation}

The introduction of MFM transforms neck fusion from fixed concatenation into channel-aware aggregation of aligned features.
The resulting representation can better reflect scale-dependent cues for sperm-head localization and discrimination between sperm, impurities, and background clutter.

\subsection{DESD Head: Detail-Enhanced Shared Detection Head}

The detection head converts multi-scale feature maps into localization and classification outputs.
In the standard YOLOv11 head, each detection scale is processed by separate convolutional branches. 
Although this design offers scale-specific flexibility, it introduces repeated head parameters and relies on conventional spatial convolutions.
SpermYOLO replaces this head with the DESD Head, which combines cross-scale parameter sharing with detail-enhanced convolution for fine-grained spatial details and localization cues.
 
\begin{figure}[t]
\centering
\includegraphics[width=0.9\textwidth]{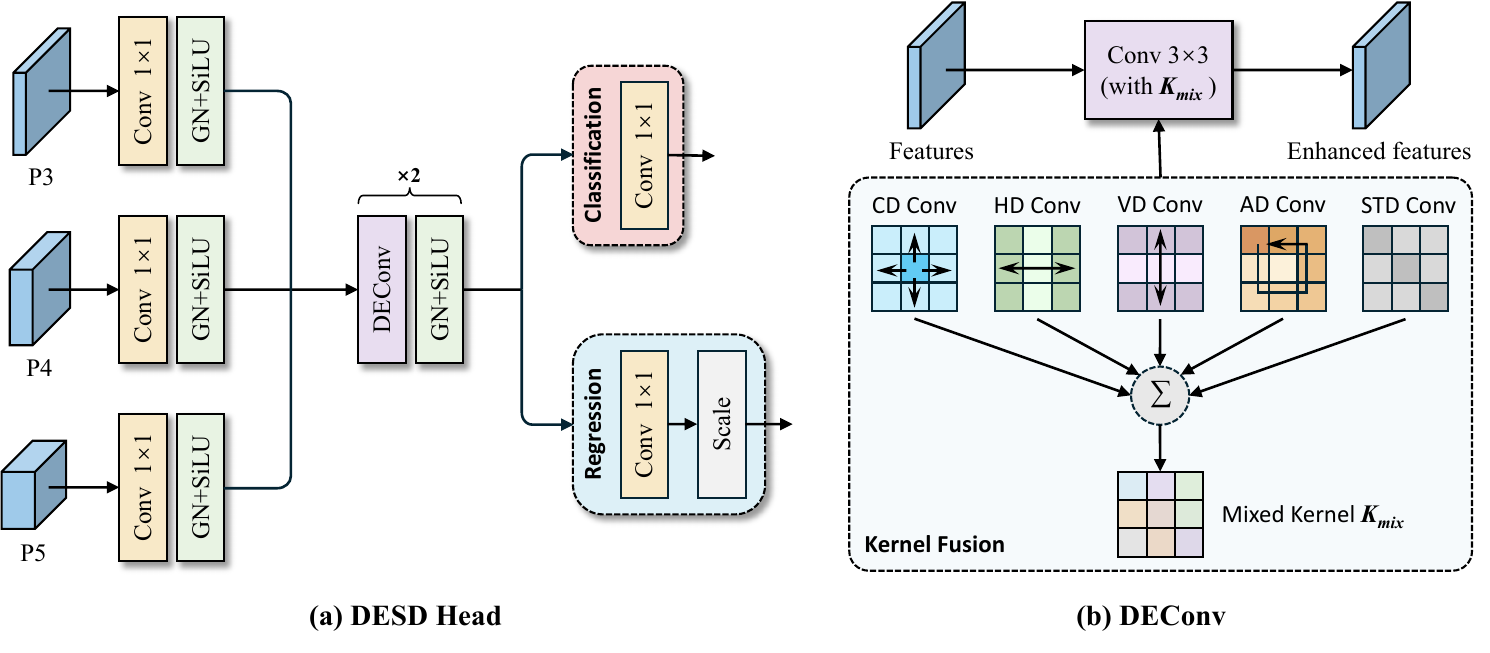}
\caption{Structure of the DESD Head}
\label{fig:desd_head}
\end{figure}

As shown in Fig.~\ref{fig:desd_head}(a), for each of the three detection scales $s\in\mathcal{S}$, the DESD Head first applies a $1\times1$ convolution with group normalization to the input feature map $X_s\in\mathbb{R}^{C_s\times H_s\times W_s}$, mapping it to a shared hidden dimension.
The scale-aligned hidden features are then processed by the same two-layer convolutional block. 
Each layer in this block is a detail-enhanced convolution (DEConv)~\citep{chen2023deanet} followed by group normalization and SiLU activation. 
The prediction output at each detection scale is therefore formulated by applying the shared regression and classification $1\times1$ projections to the processed features, with a learnable scale factor modulating the regression output as follows:
\begin{equation}
o_s =
\left[
\lambda_s g_{\mathrm{reg}}(h_s),\,
g_{\mathrm{cls}}(h_s)
\right]
\label{eq:desd_prediction}
\end{equation}
where $o_s$ denotes the prediction output at scale $s$, $h_s$ is the corresponding shared-block output, $g_{\mathrm{reg}}$ and $g_{\mathrm{cls}}$ denote the shared regression and classification projections, and $\lambda_s$ is the learnable regression scale factor.

The local detail enhancement in the shared block is introduced through DEConv.
As shown in Fig.~\ref{fig:desd_head}(b), DEConv constructs an equivalent $3\times3$ convolution from five complementary kernels: central difference convolution, horizontal difference convolution, vertical difference convolution, angular difference convolution, and standard $3\times3$ convolution.
During the forward pass, the equivalent weight and bias are obtained by summing the corresponding terms:
\begin{equation}
W_{\mathrm{DEConv}} =
W_{\mathrm{CDC}} + W_{\mathrm{HDC}} + W_{\mathrm{VDC}} + W_{\mathrm{ADC}} + W_{\mathrm{SC}}
\label{eq:deconv_weight}
\end{equation}
\begin{equation}
b_{\mathrm{DEConv}} =
b_{\mathrm{CDC}} + b_{\mathrm{HDC}} + b_{\mathrm{VDC}} + b_{\mathrm{ADC}} + b_{\mathrm{SC}}
\label{eq:deconv_bias}
\end{equation}

Through kernel fusion, DEConv incorporates directional contrast responses into the shared head, enabling it to capture local intensity variations around weak boundaries and sperm-like objects.
The DESD Head thus maintains a compact shared prediction structure while improving sensitivity to local details needed for fine-grained localization.

\section{Experimental Setup}
\label{sec:experiments}

\subsection{Datasets}

The primary experiments were performed on SVIA~\citep{chen2022svia}, a public benchmark for joint sperm--impurity detection in microscopic videos acquired with a WLJY-9000 CASA system equipped with a $20\times$ objective lens and a $20\times$ electronic eyepiece.
Its detection subset, Subset-A, contains 3,622 frames from 101 videos at $698\times528$-pixel resolution, with bounding-box annotations for over 125,000 sperm and impurity objects.
SVIA Subset-A covers semen-microscopy scenes with sparse and crowded sperm distributions, impurities of varying scales, and sperm-like distractors, making it a challenging two-class benchmark for sperm localization and sperm--impurity discrimination.
Following prior work~\citep{chen2024active}, the 101 videos were partitioned into 61 training, 20 validation, and 20 test subsets, yielding 2,146, 659, and 817 images, respectively.

Additional experiments were conducted on SDTB~\citep{zhang2025spermd} to evaluate cross-scene effectiveness in testicular-biopsy microscopy. 
SDTB contains 1,341 images and 5,548 annotated sperm instances from 15 azoospermia patients; the samples were acquired from testicular tissue sections examined under a Nikon Eclipse Ti microscope at $200\times$ magnification.
Each original image has a resolution of $1320\times1080$ pixels. Unlike SVIA, SDTB contains only the sperm class, but the targets are extremely small under complex tissue backgrounds: the reported average sperm width and height are 9.88 and 9.92 pixels, and 99.3\% of sperm instances are smaller than $16\times16$ pixels. 
The official patient-level split was used, with patients 1--7 for training, 8--10 for validation, and 11--15 for testing, giving 689, 346, and 306 images, respectively.

\subsection{Evaluation Metrics}

Average precision (AP) is used as the primary accuracy metric. For class $i$ under an intersection over union (IoU) threshold $\tau$, AP is defined as the area under the precision--recall curve:

\begin{equation}
\mathrm{AP}_{i}(\tau)=\int_{0}^{1} P_i(R;\tau)\,dR
\label{eq:ap}
\end{equation}

where $P_i(R;\tau)$ denotes the precision of class $i$ at recall $R$. The mean average precision (mAP) at threshold $\tau$ reflects the average AP over all $C$ classes:

\begin{equation}
\mathrm{mAP}_{\tau}=\frac{1}{C}\sum_{i=1}^{C}\mathrm{AP}_{i}(\tau)
\label{eq:map}
\end{equation}

We report $\mathrm{mAP}_{50}$ at an IoU threshold of 0.50 and $\mathrm{mAP}_{50:95}$ averaged over thresholds from 0.50 to 0.95 in steps of 0.05.
Precision and recall are calculated as:

\begin{equation}
P=\frac{\mathrm{TP}}{\mathrm{TP}+\mathrm{FP}},\quad
R=\frac{\mathrm{TP}}{\mathrm{TP}+\mathrm{FN}}
\label{eq:pr}
\end{equation}

Here, $\mathrm{TP}$, $\mathrm{FP}$, and $\mathrm{FN}$ represent true positives, false positives, and false negatives, respectively. The F1-score is additionally reported as the harmonic mean of precision and recall:

\begin{equation}
F1=\frac{2PR}{P+R}
\label{eq:f1}
\end{equation}

Model efficiency is evaluated in terms of trainable parameters (Params), computational complexity (GFLOPs), and inference speed (FPS), where FPS is measured as the average end-to-end processing speed on the test set, including pre-processing, inference, and post-processing. Serialized weight size is also reported when applicable.

\subsection{Implementation Details}

Experiments were conducted using an NVIDIA GeForce RTX 3090 GPU (24 GB) and an Intel Xeon Gold 6133 CPU, with Python 3.11.8, PyTorch 2.2.1, CUDA 12.1, and NumPy 1.26.4. 
All images were resized to $640\times640$ pixels for both training and evaluation. 
Under this setting, all models were trained from scratch and optimized with stochastic gradient descent (SGD). 
The batch size was set to 16 for both training and testing. 
The number of training epochs was set to 200 for SVIA and 400 for SDTB. 
Unless otherwise specified, the remaining hyperparameters followed the default configurations of the respective official implementations.

Online augmentation was introduced during training to increase image-level variation, including mosaic augmentation ($p=1.0$), random horizontal flipping ($p=0.5$), and HSV perturbation ($h=0.015$, $s=0.7$, $v=0.4$). 
Scaling and translation were set to 0.5 and 0.1, respectively.

SpermYOLO hyperparameters were fixed across both datasets. 
In C3k2-IDB, the expansion ratio was set to 4, and the dropout rate was set to 0.1. 
In D2SEM, the SEFFN expansion ratio $r$ was set to 2, and the dilated $3\times3$ DWConv used dilation $d=2$. 
For the DESD Head, the channel dimension of the shared hidden feature $h_s$ was set to 256.

\section{Results}
\label{sec:results}

In this section, we compare SpermYOLO with the baseline and analyze the contribution of each proposed module on SVIA.
We then compare it with representative generic, sperm-specific, and improved YOLO detectors, followed by cross-scene evaluation on SDTB and qualitative analysis.

\subsection{Baseline Comparison and Ablation Study}

\begin{figure}
\centering
\includegraphics[width=\textwidth]{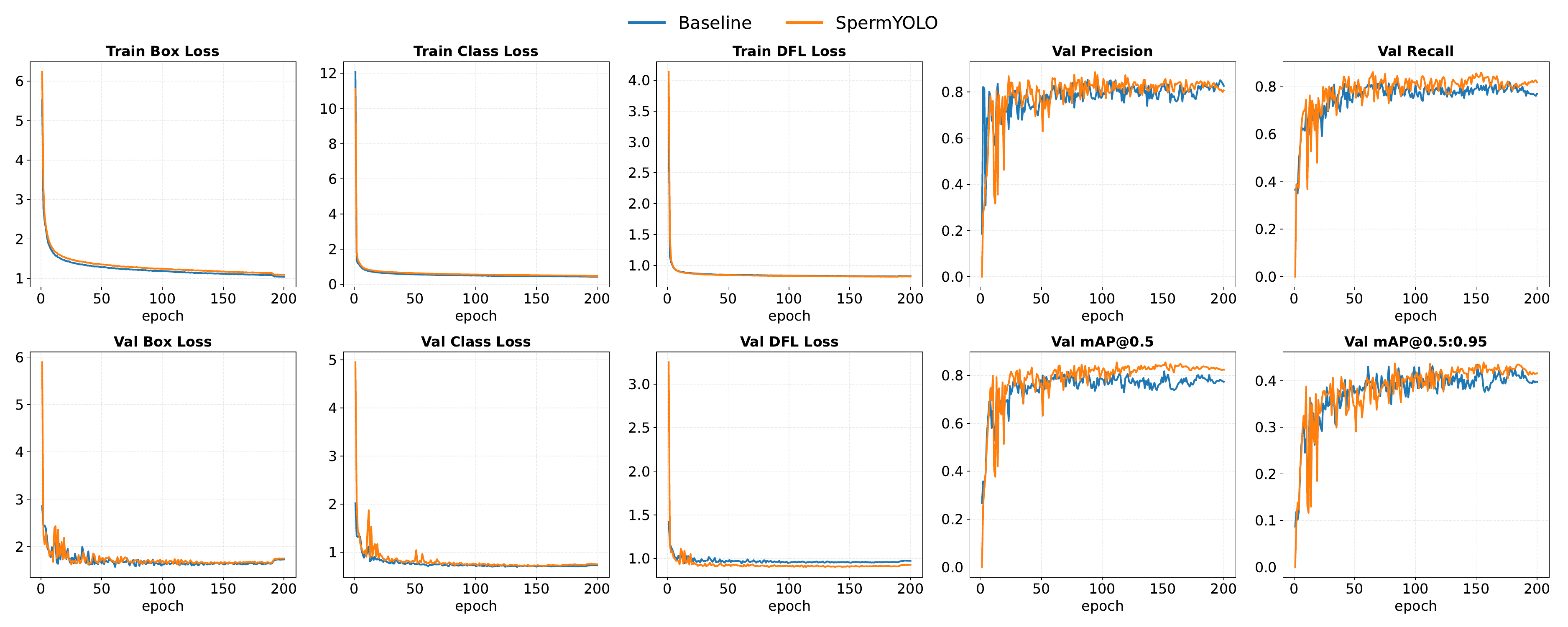}
\caption{Training and validation curves of the baseline model and SpermYOLO on the SVIA dataset.}
\label{fig:svia_training_curves}
\end{figure}

Fig.~\ref{fig:svia_training_curves} reports the training and validation curves of the baseline and SpermYOLO on SVIA under the 200-epoch schedule.
For both models, the training losses decreased rapidly during the early epochs and then stabilized, indicating convergence.
During later training, SpermYOLO attained higher validation recall, $\mathrm{mAP}_{50}$, and $\mathrm{mAP}_{50:95}$, while validation precision remained close to the baseline.
Test performance was evaluated using the checkpoint with the best validation performance for each model.

\subsubsection{Module Ablation and P-R Analysis}

To verify the effectiveness of the proposed modules, we conducted ablation experiments on SVIA, with single-factor ablation assessing their independent contributions and multi-factor ablation examining their combined effects.

\begin{table}
\caption{Ablation study of the proposed modules on the SVIA dataset.}
\label{tab:ablation}
\centering
\resizebox{\textwidth}{!}{%
\begin{tabular}{l *{14}{c}}
\toprule
& C3k2-IDB & D2SEM & MFM & DESD Head & $\mathrm{AP}_{S}$ & $\mathrm{AP}_{I}$ & $\mathrm{mAP}_{50}$ & $\mathrm{mAP}_{50:95}$ & P & R & F1 & Params & GFLOPs & Weights \\
\midrule
YOLOv11n (Baseline)  &      &       &      &            & 95.6 & 65.4 & 80.5 & 44.5 & 85.5 & 73.8 & 78.7 & 2.6M & 6.3  & 5.2MB \\
\multirow{4}{*}{Single-factor Ablation} & $\checkmark$ &       &      &            & 96.7 & 68.7 & 82.7 & 45.2 & 86.3 & 75.2 & 79.9 & 2.9M & 6.1  & 5.8MB \\
           &      & $\checkmark$ &      &            & 96.2 & 69.8 & 83.0 & 43.4 & 87.1 & 74.8 & 79.6 & 2.6M & 6.3  & 5.3MB \\
           &      &       & $\checkmark$ &            & 94.7 & 68.0 & 81.4 & 42.8 & 77.6 & 78.9 & 78.3 & 2.6M & 6.3  & 5.2MB \\
           &      &       &      & $\checkmark$ & 96.1 & 74.2 & 85.1 & 46.3 & 84.2 & 79.0 & 80.1 & 2.3M & 6.0  & 5.0MB \\
\midrule
\multirow{6}{*}{Multi-factor Ablation} & $\checkmark$ &       &      & $\checkmark$ & 96.2 & 70.5 & 83.3 & 45.7 & 79.4 & 82.0 & 80.7 & 2.6M & 5.8  & 5.6MB \\
           &      & $\checkmark$ &      & $\checkmark$ & 97.0 & 71.7 & 84.4 & 45.6 & 88.5 & 73.8 & 79.7 & 2.3M & 6.0  & 5.0MB \\
           &      &       & $\checkmark$ & $\checkmark$ & 95.1 & 68.2 & 81.7 & 43.9 & 83.0 & 76.0 & 78.9 & 2.3M & 6.0  & 5.0MB \\
           & $\checkmark$ & $\checkmark$ & $\checkmark$ &            & 96.3 & 68.6 & 82.5 & 43.1 & 85.9 & 72.6 & 78.1 & 2.9M & 6.2  & 5.9MB \\
           &      & $\checkmark$ & $\checkmark$ & $\checkmark$ & 96.2 & 77.6 & 86.9 & 47.0 & 81.9 & 80.4 & 80.9 & 2.3M & 6.1  & 5.0MB \\
           & $\checkmark$ & $\checkmark$ &      & $\checkmark$ & 95.9 & 78.7 & 87.3 & 47.6 & 79.5 & 85.4 & 82.4 & 2.6M & 5.9  & 5.7MB \\
\midrule
SpermYOLO (Ours) & $\checkmark$ & $\checkmark$ & $\checkmark$ & $\checkmark$ & 97.2 & 75.4 & 86.3 & 47.2 & 84.2 & 81.6 & 82.7 & 2.6M & 5.9 & 5.7MB \\
\bottomrule
\end{tabular}%
}
\vspace{4pt}

\end{table}

As shown in Table~\ref{tab:ablation}, all four components improve $\mathrm{mAP}_{50}$ over the baseline model in the single-factor ablations.
The DESD Head achieves the largest gain: $\mathrm{mAP}_{50}$ increases from 80.5\% to 85.1\% and $\mathrm{mAP}_{50:95}$ from 44.5\% to 46.3\%, while the parameters decrease from 2.6M to 2.3M. The improvement is particularly evident for impurities, with $\mathrm{AP}_{I}$ increasing from 65.4\% to 74.2\%. This result is consistent with the intended role of detail-enhanced convolution in strengthening weak boundary and texture cues for fine-grained prediction.
C3k2-IDB raises both class-wise and aggregate performance, increasing $\mathrm{AP}_{S}$ by 1.1 points, $\mathrm{AP}_{I}$ by 3.3 points, $\mathrm{mAP}_{50}$ by 2.2 points, and $\mathrm{mAP}_{50:95}$ by 0.7 points. These gains indicate that the inverted bottleneck with channel recalibration enhances feature discrimination while maintaining localization performance under stricter IoU thresholds.
D2SEM increases $\mathrm{AP}_{S}$ by 0.6 points, $\mathrm{AP}_{I}$ by 4.4 points, and $\mathrm{mAP}_{50}$ by 2.5 points, and attains the highest precision among the single-module variants (87.1\%).
Although it does not improve $\mathrm{mAP}_{50:95}$, the gains in class-wise AP and precision indicate that spatial--spectral enhancement mainly contributes to semantic discrimination and target recognition rather than strict localization refinement.
MFM produces a smaller $\mathrm{mAP}_{50}$ improvement (80.5\% to 81.4\%) but raises recall from 73.8\% to 78.9\%, suggesting that adaptive scale-wise fusion mainly facilitates target recovery, with its benefit becoming more evident when combined with detail-sensitive prediction and discriminative feature refinement.

The multi-factor ablations reveal interactions among the proposed modules rather than uniformly additive gains.
Among the two-factor settings, C3k2-IDB+DESD Head increases recall to 82.0\% and F1-score to 80.7\%, but decreases precision to 79.4\%, reflecting greater target recovery accompanied by more non-target responses.
D2SEM+DESD Head achieves the highest $\mathrm{AP}_{S}$ (97.0\%) and precision (88.5\%) among the two-factor variants, suggesting better sperm identification and a conservative prediction pattern with fewer false positives. 
In contrast, MFM+DESD Head produces a smaller improvement, with $\mathrm{mAP}_{50}$ of 81.7\% and $\mathrm{mAP}_{50:95}$ of 43.9\%, indicating that adaptive scale fusion alone is insufficient to fully exploit the detail-sensitive detection head.

The three-factor configurations further clarify the contribution of each retained component. 
Without C3k2-IDB, the D2SEM+MFM+DESD Head still achieves strong impurity detection, with $\mathrm{AP}_{I}$ of 77.6\%, $\mathrm{mAP}_{50}$ of 86.9\%, and $\mathrm{mAP}_{50:95}$ of 47.0\%.
This result demonstrates that spectral enhancement, adaptive fusion, and detail-enhanced prediction can jointly improve impurity-related discrimination.
However, its $\mathrm{AP}_{S}$ is 96.2\%, indicating that channel-wise recalibration remains useful for further improving sperm detection.
Without DESD Head, C3k2-IDB+D2SEM+MFM maintains $\mathrm{AP}_{S}$ of 96.3\% and precision of 85.9\%, but records lower $\mathrm{mAP}_{50}$, $\mathrm{mAP}_{50:95}$, recall, and F1-score of 82.5\%, 43.1\%, 72.6\%, and 78.1\%, respectively.
This result indicates that feature enhancement and adaptive fusion alone are insufficient and that the detail-enhanced head remains necessary to convert refined features into effective detections.
Among the three-factor variants, C3k2-IDB+D2SEM+DESD Head achieves the highest $\mathrm{mAP}_{50}$ (87.3\%) and $\mathrm{mAP}_{50:95}$ (47.6\%), showing that channel-wise recalibration, spectral semantic enhancement, and detail-enhanced prediction form the strongest three-module combination in terms of aggregate AP.
Nevertheless, its precision decreases to 79.5\%, and the $\mathrm{mAP}_{50}$ gain is driven mainly by impurity AP while $\mathrm{AP}_{S}$ drops to 95.9\%. Thus, this combination favors impurity detection and target recovery over balanced class-wise detection.

Compared with the baseline, the full SpermYOLO model increases $\mathrm{AP}_{S}$ from 95.6\% to 97.2\%, $\mathrm{AP}_{I}$ from 65.4\% to 75.4\%, $\mathrm{mAP}_{50}$ from 80.5\% to 86.3\%, and $\mathrm{mAP}_{50:95}$ from 44.5\% to 47.2\%, while retaining 2.6M parameters and reducing GFLOPs from 6.3 to 5.9.
It also improves recall from 73.8\% to 81.6\% and F1-score from 78.7\% to 82.7\%.
Relative to the strongest three-module combination, the full model has slightly lower $\mathrm{mAP}_{50}$ and $\mathrm{mAP}_{50:95}$ but higher $\mathrm{AP}_{S}$, precision, and F1-score.
Considering the practical requirement of handling sperm as the primary object, with impurity detection mainly serving to reduce sperm-like interference, SpermYOLO offers a more reliable trade-off for joint sperm--impurity detection and is therefore selected as the final configuration.

\begin{figure}
\centering
\includegraphics[width=0.8\textwidth]{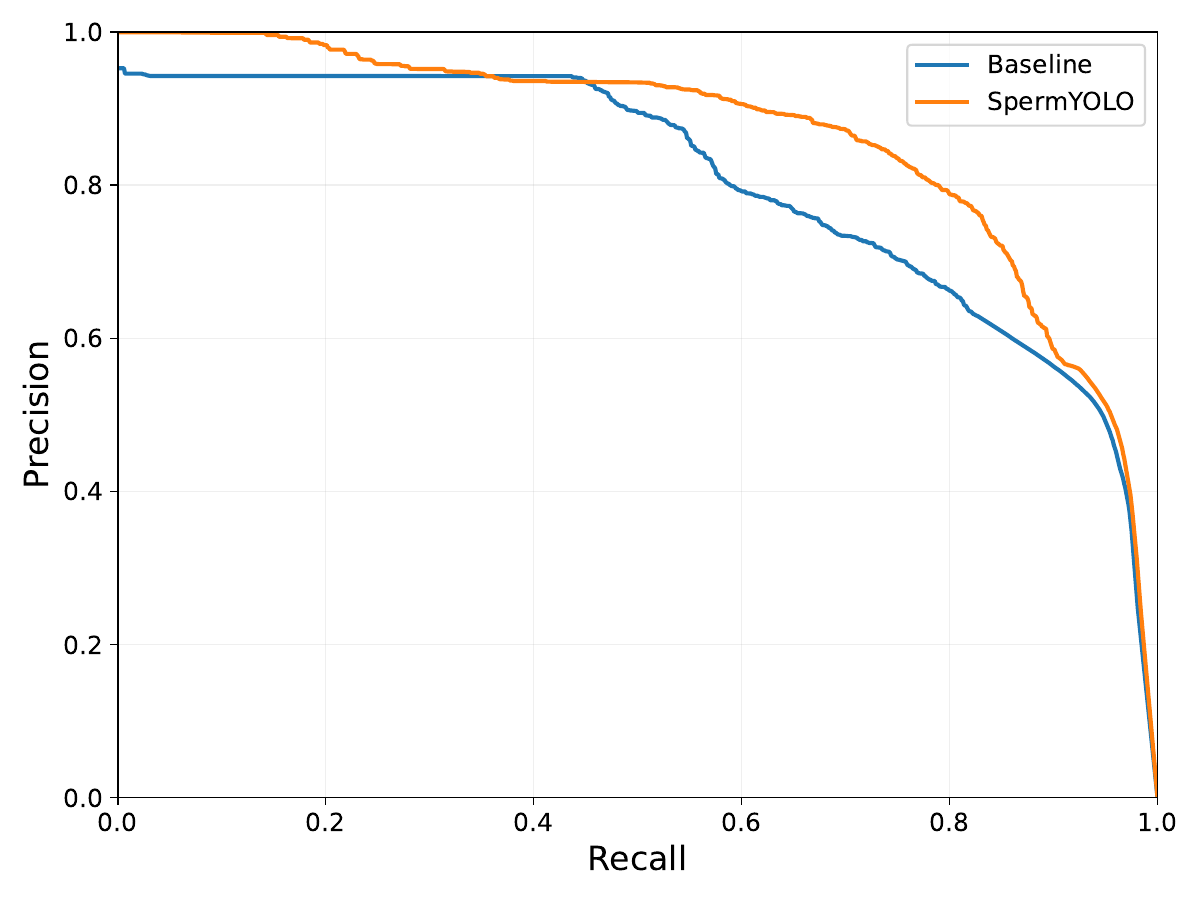}
\caption{Precision--recall curves of the baseline and SpermYOLO on the SVIA dataset.}
\label{fig:svia_pr_curve}
\end{figure}

Table~\ref{tab:ablation} also shows that SpermYOLO has slightly lower precision than the baseline on SVIA (84.2\% versus 85.5\%). 
Under the adopted evaluation protocol, precision and recall are reported at the confidence threshold that maximizes the F1 score, rather than at a fixed threshold. 
As shown in Fig.~\ref{fig:svia_pr_curve}, SpermYOLO maintains higher precision across most medium- and high-recall ranges, whereas the baseline curve declines more rapidly as recall increases. 
This curve-level comparison is consistent with the AP-based metrics, reflecting stronger performance across a wider range of confidence thresholds.

\subsubsection{Ablation Experiments for DESD Hidden Channels}

The hidden channel dimension controls the capacity of the shared representation in the DESD Head. 
This setting is important because the head must preserve fine boundary and texture information while remaining compact for efficient microscopic detection. 
If the dimension is too small, detail-sensitive features may be compressed before prediction; if it is too large, the shared head may introduce redundant computation without improving discrimination.

Table~\ref{tab:hidc} reports the sensitivity of the DESD Head to the hidden channel dimension. 
The best overall performance is obtained with the setting using 256 hidden channels, which achieves the highest $\mathrm{AP}_{S}$, $\mathrm{AP}_{I}$, $\mathrm{mAP}_{50}$, $\mathrm{mAP}_{50:95}$, precision, recall, and F1-score among the tested settings. 
Reducing the dimension to 128 gives the lowest computational cost but decreases $\mathrm{mAP}_{50:95}$ and F1-score to 44.8\% and 78.1\%, respectively, indicating that an overly narrow representation limits detail-sensitive prediction.

Increasing the hidden dimension beyond 256 does not bring further accuracy gains. 
At 512 channels, GFLOPs increase from 5.9 to 9.8, whereas $\mathrm{mAP}_{50}$ and $\mathrm{mAP}_{50:95}$ decrease to 85.6\% and 46.1\%.
This trend becomes more pronounced at 1024 hidden channels, where the parameter count and GFLOPs rise to 3.8M and 25.0, but $\mathrm{mAP}_{50}$ drops to 84.8\%. 
These results suggest that enlarging the shared head adds computation without improving the fine-grained discrimination required for compact microscopic detection. Therefore, the final SpermYOLO configuration adopts 256 hidden channels for the DESD Head.

\begin{table}
\caption{DESD Head hidden channel analysis on the SVIA dataset.}
\label{tab:hidc}
\centering
\resizebox{\textwidth}{!}{%
\begin{tabular}{c *{11}{c}}
\toprule
Hidden Channels & $\mathrm{AP}_{S}$ & $\mathrm{AP}_{I}$ & $\mathrm{mAP}_{50}$ & $\mathrm{mAP}_{50:95}$ & P & R & F1 & Params & GFLOPs & Weights \\
\midrule
128  & 96.2 & 74.3 & 85.3 & 44.8 & 78.4 & 78.0 & 78.1 & \textbf{2.5M} & \textbf{4.9} & \textbf{5.2MB} \\
\midrule
\textbf{256} & \textbf{97.2} & \textbf{75.4} & \textbf{86.3} & \textbf{47.2} & \textbf{84.2} & \textbf{81.6} & \textbf{82.7} & 2.6M & 5.9 & 5.7MB \\
\midrule
512  & 95.9 & 75.2 & 85.6 & 46.1 & 81.1 & 79.5 & 80.2 & 2.8M & 9.8 & 7.3MB \\
1024 & 95.9 & 73.7 & 84.8 & 45.3 & 82.4 & 77.3 & 79.5 & 3.8M & 25.0 & 13.6MB \\
\bottomrule
\end{tabular}%
}
\vspace{4pt}
\end{table}

\subsubsection{Ablation Experiments for C3k2-IDB Placement}

The placement of C3k2-IDB determines where channel-wise feature recalibration is introduced into the detection pipeline. 
When inserted into the backbone, C3k2-IDB mainly strengthens the representation of local visual cues, whereas insertion into the neck may improve discriminative feature selection across feature levels during multi-scale aggregation. 
Evaluating these placements separately therefore determines whether C3k2-IDB contributes mainly to feature extraction, feature fusion, or both.

Table~\ref{tab:position} examines the placement of C3k2-IDB. 
The backbone-only and neck-only settings yield similar $\mathrm{mAP}_{50}$ values of 83.3\% and 83.6\%, respectively.
The neck-only setting gives higher $\mathrm{AP}_{I}$ and $\mathrm{mAP}_{50:95}$ than the backbone-only setting but lower precision and F1-score, indicating that neck recalibration alone does not provide the most balanced detection behavior.

Applying C3k2-IDB to both the backbone and neck leads to the best overall performance. 
This setting achieves the highest $\mathrm{AP}_{S}$, $\mathrm{AP}_{I}$, $\mathrm{mAP}_{50}$, $\mathrm{mAP}_{50:95}$, precision, recall, and F1-score among the tested placements, while keeping GFLOPs slightly lower than the single-placement variants. 
These results suggest that channel-wise recalibration is useful at both stages: it benefits both feature discrimination during backbone extraction and multi-scale feature selection in the neck. 
Therefore, C3k2-IDB is deployed in both stages in the final SpermYOLO configuration.

\begin{table}
\caption{C3k2-IDB placement analysis on the SVIA dataset.}
\label{tab:position}
\centering
\resizebox{\textwidth}{!}{%
\begin{tabular}{c *{11}{c}}
\toprule
Configuration & $\mathrm{AP}_{S}$ & $\mathrm{AP}_{I}$ & $\mathrm{mAP}_{50}$ & $\mathrm{mAP}_{50:95}$ & P & R & F1 & Params & GFLOPs & Weights \\
\midrule
Backbone only   & 95.7 & 70.9 & 83.3 & 43.7 & 82.9 & 76.7 & 79.4 & \textbf{2.4M} & 6.0 & \textbf{5.2MB} \\
Neck only       & 95.7 & 71.6 & 83.6 & 45.4 & 78.9 & 78.9 & 78.7 & 2.5M & 6.0 & 5.5MB \\
\midrule
\textbf{Both} & \textbf{97.2} & \textbf{75.4} & \textbf{86.3} & \textbf{47.2} & \textbf{84.2} & \textbf{81.6} & \textbf{82.7} & 2.6M & \textbf{5.9} & 5.7MB \\
\bottomrule
\end{tabular}%
}
\vspace{4pt}
\end{table}

\subsection{Comparison with Representative Detectors on SVIA}

To evaluate SpermYOLO on SVIA~\citep{chen2022svia}, we compared it with representative detectors from three categories. 
The general-purpose category includes Faster R-CNN~\citep{ren2017faster}, SSD~\citep{liu2016ssd}, YOLO-series models~\citep{li2022yolov6,varghese2024yolov8,wang2024yolov10,khanam2024yolov11,tian2025yolov12,lei2025yolov13}, and RT-DETR~\citep{zhao2024detrs}, covering two-stage, one-stage, and transformer-based paradigms. 
The sperm-specific category includes SpermDet~\citep{zhang2025spermd}, DeepSperm~\citep{hidayatullah2021deepsperm}, and ACTIVE~\citep{chen2024active}, which are designed for sperm detection or joint sperm--impurity analysis in microscopic images and videos. 
The YOLO-improved category consists of LSM-YOLO~\citep{yu2024lsmyolo} and CST-YOLO~\citep{kang2023cstyolo} as biomedical-oriented enhancements, with CST-YOLO targeting small objects in blood-cell imagery. It further includes MAF-YOLO~\citep{yang2024mafyolo} and YOLOv5n-SPD~\citep{sunkara2022spdconv}, which are designed for small-object, multi-scale, or low-resolution scenarios closely matching the challenges of dense sperm microscopy images.

\begin{table}[t]
\caption{Comparison with representative detectors on the SVIA dataset.}
\label{tab:svia}
\centering
\setlength{\tabcolsep}{2.5pt}
\scriptsize
\resizebox{\textwidth}{!}{%
\begin{tabular}{llccccccccccc}
\toprule
Category & Model & $\mathrm{AP}_{S}$ & $\mathrm{AP}_{I}$ & $\mathrm{mAP}_{50}$ & $\mathrm{mAP}_{50:95}$ & P & R & F1 & Params & GFLOPs & FPS & Weights \\
\midrule
\multirow{10}{*}{Generic}
& Faster R-CNN~\citep{ren2017faster} & 93.2 & 49.4 & 71.3 & 37.5 & 71.2 & 68.3 & 69.4 & 41.1M & 167.5 & 9.7 & 158.0MB \\
& SSD~\citep{liu2016ssd}             & 82.5 & 46.4 & 64.5 & 23.0 & 75.7 & 58.7 & 65.0 & 23.9M & 30.5 & 43.2 & 91.1MB \\
& YOLOv5n                            & 94.2 & 64.8 & 79.5 & 42.4 & 80.3 & 73.2 & 76.3 & 2.5M & 7.1 & 98.0 & 5.0MB \\
& YOLOv6n~\citep{li2022yolov6}        & 91.6 & 62.1 & 76.9 & 34.3 & 79.8 & 74.0 & 76.5 & 4.2M & 11.8 & 89.0 & 8.3MB \\
& YOLOv8n~\citep{varghese2024yolov8}  & 95.4 & 66.6 & 81.0 & 44.5 & 82.4 & 73.0 & 76.7 & 3.0M & 8.1 & 96.0 & 6.0MB \\
& YOLOv10n~\citep{wang2024yolov10}    & 95.3 & 67.9 & 81.6 & 41.4 & 76.9 & 78.2 & 77.4 & 2.3M & 6.5 & 161.0 & 5.5MB \\
& YOLOv11n~\citep{khanam2024yolov11}  & 95.6 & 65.4 & 80.5 & 44.5 & \textbf{85.5} & 73.8 & 78.7 & 2.6M & 6.3 & 94.4 & 5.2MB \\
& YOLOv12n~\citep{tian2025yolov12}    & 95.6 & 54.8 & 75.2 & 40.8 & 85.1 & 68.7 & 74.9 & 2.6M & 6.3 & 82.0 & 5.3MB \\
& YOLOv13n~\citep{lei2025yolov13}     & 94.9 & 59.3 & 77.1 & 38.4 & 77.0 & 72.1 & 74.0 & 2.4M & 6.2 & 104.0 & 5.2MB \\
& RT-DETR-l~\citep{zhao2024detrs}      & 91.0 & 43.7 & 67.3 & 30.0 & 65.8 & 70.9 & 68.2 & 32.0M & 103.4 & 65.0 & 63.1MB \\
\midrule
\multirow{3}{*}{Sperm-specific}
& SpermDet~\citep{zhang2025spermd}             & 95.6 & 69.3 & 82.4 & 44.9 & 75.9 & 82.4 & 79.0 & 17.4M & 99.9 & 64.0 & 33.6MB \\
& DeepSperm~\citep{hidayatullah2021deepsperm} & 94.3 & 49.0 & 71.6 & 36.3 & 69.2 & 70.3 & 68.6 & 4.2M & 70.8 & 32.1 & 16.0MB \\
& ACTIVE~\citep{chen2024active}                & 96.8 & 62.7 & 79.7 & 36.5 & 80.5 & 77.8 & 79.0 & 124.4M & 49.1 & 26.9 & 475.6MB \\
\midrule
\multirow{4}{*}{YOLO-improved}
& LSM-YOLO~\citep{yu2024lsmyolo}      & 95.7 & 61.2 & 78.5 & 43.5 & 71.5 & 77.7 & 74.5 & 2.9M & 12.4 & 123.0 & 6.0MB \\
& CST-YOLO~\citep{kang2023cstyolo}    & 95.7 & 66.3 & 81.0 & 40.5 & 80.4 & 76.0 & 77.7 & 47.5M & 235.4 & 191.0 & 92.4MB \\
& YOLOv5n-SPD~\citep{sunkara2022spdconv} & 95.5 & 73.0 & 84.3 & 41.6 & 84.5 & 75.6 & 79.6 & \textbf{2.1M} & 8.5 & 141.0 & \textbf{4.4MB} \\
& MAF-YOLO~\citep{yang2024mafyolo}    & 95.7 & 62.6 & 79.2 & 39.3 & 71.6 & \textbf{85.1} & 77.6 & 3.7M & 10.3 & \textbf{220.0} & 8.0MB \\
\midrule
Ours & SpermYOLO & \textbf{97.2} & \textbf{75.4} & \textbf{86.3} & \textbf{47.2} & 84.2 & 81.6 & \textbf{82.7} & 2.6M & \textbf{5.9} & 185.0 & 5.7MB \\
\bottomrule
\end{tabular}%
}
\end{table}

As shown in Fig.~\ref{fig:sperm_vs_impurity} and Table~\ref{tab:svia}, SpermYOLO achieves the highest AP for both sperm and impurities. 
For sperm detection, SpermYOLO obtains an AP of 97.2\%, outperforming the strongest competing method, ACTIVE, by 0.4 points. 
For impurity detection, SpermYOLO achieves an AP of 75.4\%, exceeding the second-best method, YOLOv5n-SPD, by 2.4 points. 
This improvement is notable because impurity detection is more challenging on SVIA. 
Several competing methods maintain high sperm AP but show a clear performance drop on impurities, such as ACTIVE (96.8\% $\mathrm{AP}_{S}$ versus 62.7\% $\mathrm{AP}_{I}$), SpermDet (95.6\% $\mathrm{AP}_{S}$ versus 69.3\% $\mathrm{AP}_{I}$), and CST-YOLO (95.7\% $\mathrm{AP}_{S}$ versus 66.3\% $\mathrm{AP}_{I}$). 
In contrast, SpermYOLO improves both impurity recognition and sperm detection, indicating stronger discrimination between sperm cells and sperm-like distractors.

The aggregate metrics show the same trend. 
SpermYOLO obtains the highest $\mathrm{mAP}_{50}$, $\mathrm{mAP}_{50:95}$, and F1-score among all evaluated detectors. 
Compared with YOLOv5n-SPD, the strongest competing method in $\mathrm{mAP}_{50}$, SpermYOLO improves $\mathrm{mAP}_{50}$ by 2.0 points and $\mathrm{mAP}_{50:95}$ by 5.6 points. 
Compared with SpermDet, the strongest sperm-specific model in $\mathrm{mAP}_{50}$, the corresponding gains are 3.9 and 2.3 points.
The gains in both metrics indicate that SpermYOLO improves detection at the permissive threshold while maintaining stronger localization under stricter criteria.

Although several models achieve higher precision or recall individually, their F1-scores remain lower than that of SpermYOLO. 
YOLOv11n, YOLOv12n, and YOLOv5n-SPD report slightly higher precision, while MAF-YOLO and SpermDet obtain higher recall. 
In particular, MAF-YOLO achieves the highest recall (85.1\%), but substantially lower precision (71.6\%) and $\mathrm{mAP}_{50}$ (79.2\%), suggesting that its higher target recovery is accompanied by more false positives. 
SpermYOLO nevertheless achieves the highest F1-score, showing a more balanced precision--recall trade-off for dense microscopy, where both missed sperm and false impurity responses can affect downstream analysis.

In terms of efficiency, SpermYOLO retains a compact model scale of 2.6M parameters and 5.7MB weights, with the lowest computational cost among all compared models at 5.9 GFLOPs. 
Although its FPS (185) is lower than that of MAF-YOLO (220) and CST-YOLO (191), it is still sufficient for real-time automated analysis. 
Considering its accuracy gains and compact computational profile, SpermYOLO provides a more favorable accuracy--efficiency trade-off, delivering the best overall detection performance without relying on a large model or heavy computational cost.

\subsection{Cross-Scene Evaluation on SDTB}

To assess SpermYOLO in a distinct sperm microscopy setting, we evaluated it on the SDTB~\citep{zhang2025spermd} dataset. 
Unlike SVIA, which comprises semen-smear microscopy images with annotations for both sperm and impurities, SDTB consists of testicular biopsy images with sperm annotations only.
These differences in imaging source and annotation scope provide a cross-scene test of whether the proposed architectural adaptations remain effective beyond the primary semen-smear setting.

As shown in Table~\ref{tab:sdtb}, several general-purpose detectors perform poorly on SDTB. 
Faster R-CNN, SSD, and RT-DETR-l achieve $\mathrm{mAP}_{50}$ values of 14.4\%, 12.4\%, and 37.3\%, respectively, indicating limited adaptation to testicular-biopsy microscopy.
Performance also varies considerably among YOLO-based detectors.
YOLOv6n achieves the lowest $\mathrm{mAP}_{50}$ of 54.5\%, while YOLOv5n-SPD, despite its competitive performance on SVIA, records the second-lowest $\mathrm{mAP}_{50}$ among the evaluated YOLO models.
In contrast, SpermYOLO obtains the highest $\mathrm{mAP}_{50}$ (74.8\%) and $\mathrm{mAP}_{50:95}$ (31.2\%) among all evaluated detectors.
Compared with YOLOv8n, the strongest competing YOLO model on SDTB, SpermYOLO improves $\mathrm{mAP}_{50}$ and $\mathrm{mAP}_{50:95}$ by 3.0 and 1.3 points, respectively.

Sperm-specific detectors remain competitive on SDTB, with ACTIVE and SpermDet achieving $\mathrm{mAP}_{50}$ values of 74.0\% and 73.1\%, respectively. 
Compared with ACTIVE, the strongest competing method in $\mathrm{mAP}_{50}$, SpermYOLO improves $\mathrm{mAP}_{50}$ by 0.8 points and $\mathrm{mAP}_{50:95}$ by 4.2 points. Relative to SpermDet, the second-best method in $\mathrm{mAP}_{50:95}$, the corresponding gains are 1.7 and 1.2 points, respectively.
Since sperm instances in SDTB are extremely small and embedded in complex tissue backgrounds, limited visual evidence and interference from surrounding structures complicate their identification, while minor localization errors can result in substantial IoU variation.
Accordingly, the higher $\mathrm{mAP}_{50}$ reflects SpermYOLO's more reliable target identification under complex tissue backgrounds, whereas the higher $\mathrm{mAP}_{50:95}$ reflects improved fine-grained localization.

The precision--recall results provide a complementary view.
ACTIVE obtains the highest recall of 67.8\%, followed by DeepSperm and LSM-YOLO, both at 67.5\%.
Although SpermYOLO has slightly lower recall at 67.1\%, it records the highest precision (76.3\%) and F1-score (71.4\%), providing the strongest balance between sperm recovery and false-positive suppression on SDTB.

Overall, the SDTB evaluation shows that SpermYOLO maintains strong detection and localization under the markedly different target-scale and background characteristics of testicular-biopsy microscopy. 
Its leading AP metrics and favorable precision--recall balance further support the effectiveness of the coordinated architectural adaptations in this distinct sperm microscopy scenario.

\begin{table}[t]
    \caption{Comparison with representative detectors on the SDTB dataset.}
    \label{tab:sdtb}
    \centering
    \setlength{\tabcolsep}{4pt}
    \scriptsize
    \resizebox{0.85\textwidth}{!}{%
    \begin{tabular}{llccccc}
    \toprule
    Category & Model & $\mathrm{mAP}_{50}$ & $\mathrm{mAP}_{50:95}$ & P & R & F1 \\
    \midrule
    \multirow{10}{*}{Generic}
    & Faster R-CNN~\citep{ren2017faster} & 14.4 & 4.7  & 17.2 & 43.8 & 24.7 \\
    & SSD~\citep{liu2016ssd}             & 12.4 & 3.5  & 26.9 & 15.7 & 19.8 \\
    & YOLOv5n                            & 67.8 & 28.5 & 71.5 & 60.4 & 65.5 \\
    & YOLOv6n~\citep{li2022yolov6}        & 54.5 & 22.0 & 53.9 & 57.9 & 55.8 \\
    & YOLOv8n~\citep{varghese2024yolov8}  & 71.8 & 29.9 & 74.5 & 64.7 & 69.2 \\
    & YOLOv10n~\citep{wang2024yolov10}    & 70.9 & 29.4 & 73.3 & 63.1 & 67.8 \\
    & YOLOv11n~\citep{khanam2024yolov11}  & 68.7 & 28.3 & 71.9 & 59.7 & 65.3 \\
    & YOLOv12n~\citep{tian2025yolov12}    & 70.6 & 29.6 & 73.6 & 62.8 & 67.7 \\
    & YOLOv13n~\citep{lei2025yolov13}     & 67.1 & 28.1 & 69.0 & 61.1 & 64.8 \\
    & RT-DETR-l~\citep{zhao2024detrs}    & 37.3 & 15.3 & 42.6 & 44.6 & 43.6 \\
    \midrule
    \multirow{3}{*}{Sperm-specific}
    & DeepSperm~\citep{hidayatullah2021deepsperm} & 72.5 & 27.9 & 73.8 & 67.5 & 70.5 \\
    & SpermDet~\citep{zhang2025spermd}            & 73.1 & 30.0 & 73.7 & 66.6 & 70.0 \\
    & ACTIVE~\citep{chen2024active}               & 74.0 & 27.0 & 74.4 & \textbf{67.8} & 70.9 \\
    \midrule
    \multirow{4}{*}{YOLO-improved}
    & CST-YOLO~\citep{kang2023cstyolo}    & 69.9 & 29.0 & 69.9 & 64.8 & 67.3 \\
    & LSM-YOLO~\citep{yu2024lsmyolo}      & 65.3 & 25.8 & 72.3 & 67.5 & 69.8 \\
    & YOLOv5n-SPD~\citep{sunkara2022spdconv} & 61.0 & 24.6 & 58.5 & 59.5 & 59.0 \\
    & MAF-YOLO~\citep{yang2024mafyolo}    & 69.5 & 28.9 & 71.1 & 62.9 & 66.8 \\
    \midrule
    Ours & SpermYOLO & \textbf{74.8} & \textbf{31.2} & \textbf{76.3} & 67.1 & \textbf{71.4} \\
    \bottomrule
    \end{tabular}
    }
    \end{table}

\subsection{Qualitative Analysis}

\subsubsection{Detection Result Visualization}

To further examine detection behavior, we conducted a qualitative comparison between the baseline model and SpermYOLO on representative SVIA images, complementing the quantitative evaluation.
Correct detections, false positives, and missed targets were identified based on ground-truth annotations to illustrate differences in target recovery and false-positive suppression.

\begin{figure}[!htbp]
\centering
\includegraphics[width=\textwidth]{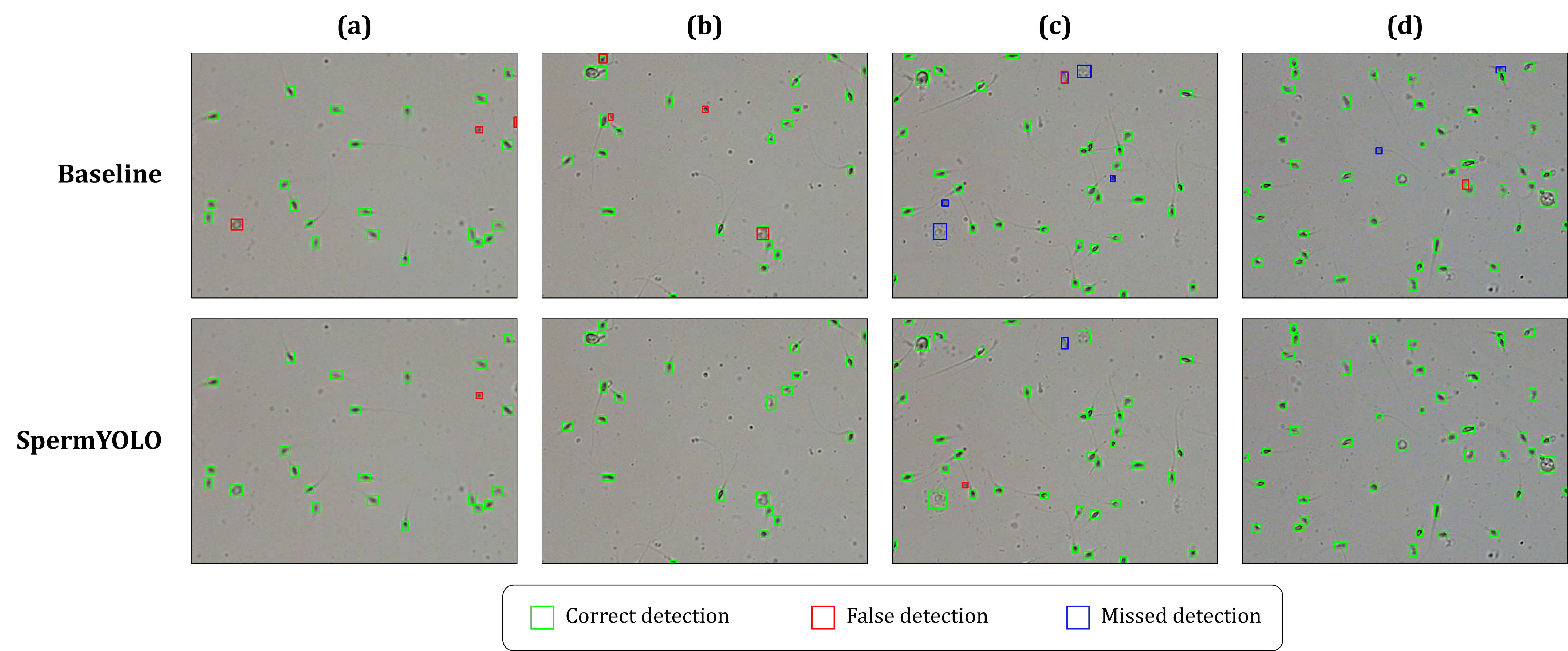}
\caption{Qualitative detection comparison between the baseline and SpermYOLO on representative SVIA images. The upper row shows baseline predictions, and the lower row shows SpermYOLO predictions on the same images. Green, red, and blue boxes denote correct detections, false-positive detections, and missed targets, respectively.}
\label{fig:vis_detection}
\end{figure}

Fig.~\ref{fig:vis_detection} shows the baseline predictions in the upper row and the corresponding SpermYOLO predictions in the lower row.
The four examples cover typical challenging cases in sperm microscopy.
Fig.~\ref{fig:vis_detection}(a) and (b) show relatively sparse scenes containing small background structures and ambiguous sperm-like particles.
In Fig.~\ref{fig:vis_detection}(a), the baseline produces three false detections, including false-positive sperm detections arising from background structures and one impurity misclassified as sperm. 
In Fig.~\ref{fig:vis_detection}(b), the baseline produces four false positives, including redundant responses around a correctly detected impurity region and responses to sperm-like background structures.
In contrast, SpermYOLO reduces or avoids most of these errors, suggesting that it is less sensitive to isolated sperm-like background particles and better distinguishes true sperm or impurity targets from visually similar non-target structures.

Fig.~\ref{fig:vis_detection}(c) presents a more complex mixed scene with dense targets, impurities, and background interference.
The baseline misses several sperm and impurity targets while also misclassifying one impurity as sperm.
Although SpermYOLO still produces one false-positive detection and misses one impurity target, it substantially reduces the number of missed detections.
In the crowded sperm-dominant scene shown in Fig.~\ref{fig:vis_detection}(d), many small sperm heads appear close together with low local contrast.
The baseline produces one false-positive sperm detection and misses two sperm, whereas SpermYOLO shows no obvious false-positive or missed detections.
These dense-scene results suggest improved recovery of crowded small targets under sperm--impurity coexistence.

Overall, the qualitative results show that SpermYOLO improves detection behavior mainly in two aspects: suppressing false-positive responses caused by sperm-like background structures and recovering more small or low-contrast sperm targets in dense microscopic scenes.

\subsubsection{Attention Heatmap Analysis}

\begin{figure}[!htbp]
\centering
\includegraphics[width=0.82\textwidth]{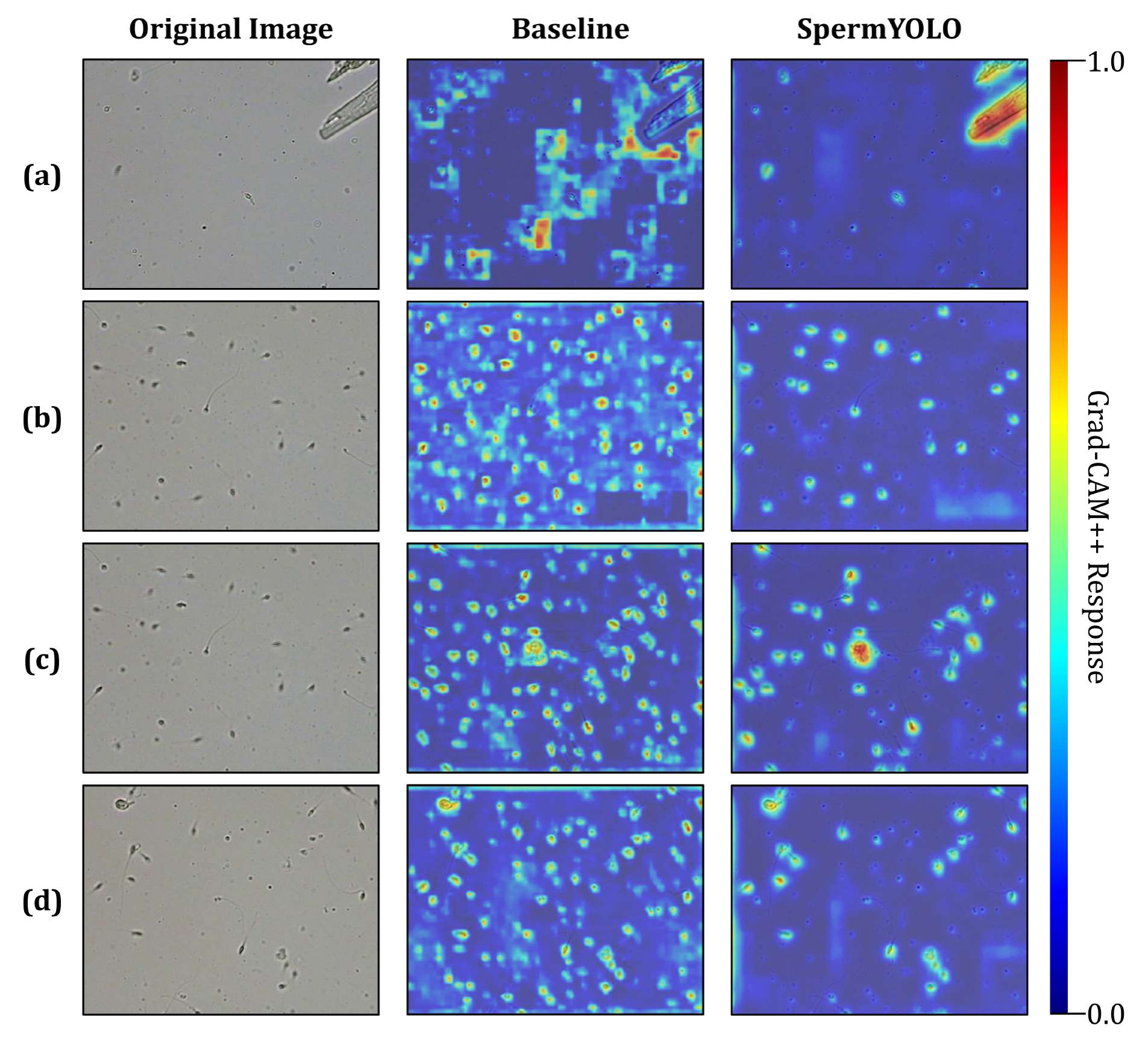}
\caption{Grad-CAM++ heatmap analysis of the baseline and SpermYOLO. From left to right, each row shows the original image, the baseline heatmap, and the SpermYOLO heatmap; warmer colors indicate stronger response.}
\label{fig:heatmap}
\end{figure}

To further characterize the spatial feature response patterns of the proposed method, Grad-CAM++~\citep{chattopadhyay2018gradcampp} was applied to the final feature maps preceding prediction at multiple detection scales for each input image. The resulting responses were then upsampled to the original input resolution and fused to generate a final heatmap, as shown in Fig.~\ref{fig:heatmap}.

The baseline produces relatively broad and scattered activations across the image. 
In Fig.~\ref{fig:heatmap}(a), block-like and grid-like high-response regions appear in non-target areas together with responses to background noise. 
For the large impurity-like structure in the upper-right region, the baseline response is confined to only a small portion with moderate intensity.
In Fig.~\ref{fig:heatmap}(b), dense point-like responses cover small background particles, circular noise-like spots, and texture-like regions. 
Large non-target regions, especially in the lower-middle area, are also covered by diffuse patch-like activations, where low-to-moderate responses are mixed with high-intensity spots, resulting in weak separation between target-related structures and ordinary background.
Similar point-like and diffuse background responses persist in Fig.~\ref{fig:heatmap}(c) and (d). 
These observations indicate that the baseline is sensitive to background noise and tends to produce spatially dispersed responses in complex microscopic scenes.

In contrast, SpermYOLO produces more compact and selective activation maps.
In Fig.~\ref{fig:heatmap}(a), the high-response regions are concentrated around salient microscopic structures, while the diffuse block-like background responses observed in the baseline are reduced.
Across Fig.~\ref{fig:heatmap}(b)--(d), SpermYOLO suppresses many scattered activations and diffuse background responses, retaining more localized responses around compact microscopic structures.
This indicates that SpermYOLO forms a more focused response pattern and is less affected by background particles and texture-like interference.

Overall, compared with the baseline, SpermYOLO shows weaker background activation and more concentrated responses around relevant microscopic structures, consistent with the quantitative improvements on SVIA.

\clearpage

\section{Conclusion}
\label{sec:conclusion}

In this paper, we present SpermYOLO, a coordinated and compact YOLOv11-derived framework for joint sperm--impurity detection in microscopic images. 
The proposed model introduces four architectural improvements targeting key challenges in sperm microscopy detection. 
C3k2-IDB enhances channel-wise discriminative feature extraction through an inverted bottleneck with lightweight channel recalibration. 
D2SEM strengthens high-level semantic representations by incorporating spatial--spectral enhancement into deep feature processing. 
MFM enables adaptive multi-scale feature fusion in the neck, and the DESD Head improves detail-sensitive prediction through shared detail-enhanced convolution. 
Together, these components provide complementary adaptations for detecting small, low-contrast sperm targets and sperm-like impurities.

Comprehensive experiments on the SVIA and SDTB datasets demonstrate the effectiveness of SpermYOLO.
On the SVIA~\citep{chen2022svia} benchmark, SpermYOLO achieves strong joint sperm--impurity detection performance, attaining 97.2\% sperm AP and 75.4\% impurity AP and outperforming generic detectors, dedicated sperm detection models, and improved YOLO variants. 
Compared with the baseline model, SpermYOLO improves joint sperm--impurity detection accuracy, especially for impurity recognition, while maintaining a compact model scale.
Cross-scene evaluation on the SDTB~\citep{zhang2025spermd} benchmark further shows that SpermYOLO remains effective for sperm detection in testicular-biopsy microscopy with extremely small targets and complex tissue backgrounds. 
Ablation studies support the contribution of the proposed modules and clarify the balanced performance of the final configuration, while qualitative analyses show improved detection behavior and more focused feature response patterns.

Several limitations of this work should be acknowledged. 
First, the evaluation has been conducted in two representative sperm imaging scenarios, namely semen smear and testicular-biopsy microscopy; performance under other sample preparation and imaging conditions, such as different staining protocols, phase-contrast microscopy, or fluorescence imaging, remains to be validated.
Second, the present study focuses on sperm detection, while subsequent motion analysis, including sperm tracking and motility-related parameter estimation, was not investigated.

Future work will validate and adapt SpermYOLO under more diverse sperm microscopy conditions, and integrate the detection framework with sperm tracking and motility analysis for comprehensive computer-aided semen assessment.
In addition, the transferability of the proposed architectural design principles to other medical small-object detection tasks will be further explored.

\section*{CRediT authorship contribution statement}
Shengqi Chen: Conceptualization, Methodology, Formal analysis, Writing -- original draft, Writing -- review \& editing, Visualization, Project administration.
Zilin Wang: Software, Methodology, Investigation, Validation, Writing -- original draft.
Xingyu Pan: Software, Investigation, Visualization, Writing -- review \& editing.
Wenting Yu: Investigation, Resources, Data curation.
Pengchao Deng: Software, Methodology, Visualization, Resources, Data curation.
Guohua Wu: Conceptualization, Funding acquisition, Project administration, Supervision.

\section*{Declaration of competing interests}
The authors declare that they have no known competing financial interests or personal relationships that could have appeared to influence the work reported in this paper.

\section*{Acknowledgements}
This work was supported by the National Natural Science Foundation of China [grant number 62475020].

\section*{Data availability}
The data that support the findings of this study are openly available on GitHub at \url{https://github.com/ElvisChen7/SpermYOLO}.

\bibliographystyle{elsarticle-num}
\bibliography{references}

\end{document}